\documentclass{article}

\usepackage{colortbl}

    \usepackage[preprint]{neurips_2025}

\usepackage[utf8]{inputenc} % allow utf-8 input
\usepackage[T1]{fontenc}    % use 8-bit T1 fonts
\usepackage{hyperref}       % hyperlinks
\usepackage{url}            % simple URL typesetting
\usepackage{booktabs}       % professional-quality tables
\usepackage{amsfonts}       % blackboard math symbols
\usepackage{nicefrac}       % compact symbols for 1/2, etc.
\usepackage{microtype}      % microtypography
\usepackage{xcolor}         % colors

\usepackage{graphicx}  % resizebox
\usepackage{threeparttable}  % footnote
\usepackage{makecell} 
\usepackage{caption}

\usepackage{tabularx} % for 'tabularx' env. and 'X' col. type
\usepackage{ragged2e} % for \RaggedRight macro
\usepackage{booktabs} % for \toprule, \midrule etc macros
\usepackage{rotating}

\usepackage{graphicx}
\usepackage{wrapfig}
\usepackage{mathrsfs}
\usepackage{amssymb}
\usepackage{algorithm}
\usepackage{algorithmic}
\usepackage{newtxtext}
\usepackage[algo2e, ruled]{algorithm2e}
\usepackage{multirow}
\usepackage{amsmath}
\usepackage{cleveref}
\definecolor{linecolor}{rgb}{0.85, 0.98, 0.8}
\newcommand{\method}[1]{\textsc{Dynamic Base Model Shift}}
\newcommand{\methodshort}[1]{\textsc{DBMS}}
\title{Dynamic Base Model Shift for Delta Compression}
\SetKwComment{Comment}{$\triangleright$\ }{}

\author{%
    \textbf{Chenyu Huang}$^{1\dagger}$\quad
    \textbf{Peng Ye}$^{1,2,3\dagger}$\quad
    Shenghe Zheng$^{2, 4}$\\
    \textbf{Xiaohui Wang}$^1$\quad
    \textbf{Lei Bai}$^{2}$ \quad
    \textbf{Tao Chen}$^{1}$\thanks{Corresponding Author(eetchen@fudan.edu.cn).~~~$^\dagger$Equal Contribution.} \quad
    \textbf{Wanli Ouyang}$^{2,3}$ 
    \\ 
  $^1$ Fudan University \quad $^2$ Shanghai AI Laboratory \\ $^3$ The Chinese University of Hong Kong \quad $^4$ Harbin Institute of Technology \\
  {\tt\small cyhuang24@m.fudan.edu.cn}
}
\begin{document}

\maketitle

\begin{abstract}

Transformer-based models with the pretrain-finetune paradigm bring about significant progress, along with the heavy storage and deployment costs of finetuned models on multiple tasks.
Delta compression attempts to lower the costs by reducing the redundancy of delta parameters (i.e., the difference between the finetuned and pre-trained model weights) through pruning or quantization.
However, existing methods by default employ the pretrained model as the base model and compress the delta parameters for every task, which may causes significant performance degradation,
% compared to uncompressed finetuned models, 
especially when the compression rate is extremely high.
To tackle this issue, we investigate the impact of different base models on the performance of delta compression and find that the pre-trained base model can hardly be optimal.
To this end, we propose \textsc{Dynamic Base Model Shift} (\textsc{DBMS}), which dynamically adapts the base model to the target task before performing delta compression.
% to reduce performance loss.
Specifically, we adjust two parameters, which respectively determine the magnitude of the base model shift and the overall scale of delta compression, to boost the compression performance on each task.
Through the low-cost learning of these two parameters, our \textsc{DBMS} can maintain most of the finetuned model's performance even under an extremely high compression ratio setting, significantly surpassing existing methods.
Moreover, our \textsc{DBMS} is orthogonal and can be integrated with a variety of other methods. It has been evaluated across different types of models, including language models, vision transformers, and multi-modal models.

\end{abstract}

\section{Introduction}
\label{sec:introduction}

As transformer-based models become mainstream~\cite{vaswani2017attention, devlin2018bert, achiam2023gpt, liu2024deepseek} along with the pretrain-finetune paradigm, the model size and number are becoming increasingly larger, thus storing models finetuned on different downstream tasks is getting much more expensive.
To this end, different techniques have been proposed to reduce storage and deployment costs.
A representative technique is model merging, which means fusing model weights from different models~\cite{wortsman2022model, RegMean, TIESMerging}. 
Though showing impressive performance and wide applicability~\cite{yang2024model}, model merging still suffers from significant performance degradation when the number or difficulty of target tasks increases~\cite{huang2024emrmerging, lu2024twin, zheng2024free, tang2024merging, tang2025data, kim2025task}.
Therefore, delta compression, which compresses the delta parameters or task vectors~\cite{Task_Arithmetic}, i.e., the difference between the finetuned and the pre-trained model, has been recently proposed~\cite{liu2024bitdelta,ping2024deltacome}.
Similar to model merging, delta compression also utilizes the extremely redundant properties of delta parameters of finetuned models, but differently, delta compression is not affected by weight interference, thus achieving more stable performance.

% Through delta compression, the storage costs of delta parameters can be sharply reduced.
Currently, delta compression methods can be roughly divided into pruning-based methods and quantization-based methods.
Pruning-based methods eliminate most elements within delta parameters randomly~\cite{yu2023language, deng2024dare} or based on magnitude~\cite{TIESMerging, deng2024dare}.
Quantization-based methods apply post-training quantization (PTQ)~\cite{frantar2022gptq} to delta parameters to quantize them to 1-bit~\cite{liu2024bitdelta} or equivalently 1-bit~\cite{ping2024deltacome, huang2025seeing} through mixed-precision quantization.
Both pruning-based and quantization-based methods can compress the delta parameters, thus significantly reducing the storage costs.
However, existing methods apply the pre-trained model as the base model for delta compression by default, which may not be optimal.
As shown in the left part of Fig.~\ref{fig:moti}, there is much room for lowering the distance between the base model and different finetuned models.
In comparison, after applying our \method{}, the distance could be significantly reduced, as shown in Fig.~\ref{fig:moti}(c).

To resolve the mentioned issue, we rethink and analyze the existing delta compression paradigm. 
When handling $N$ downstream tasks $\left[T_1, T_2, ..,T_N\right]$, 
% instead of storing 
which includes $N$ model weights $\left[\theta_1, \theta_2, ..,\theta_N\right]$ finetuned from the same pre-trained model $\theta_{pre}$, 
% we store $\theta_{pre}$ and $\left[\delta_1, \delta_2,..,\delta_N\right]$, where $\delta_t$ is model $\theta_t$'s compressed delta parameter 
% $\Delta_t = \theta_t-\theta_{pre}$,
the goal of existing methods is to compress the differences between finetuned models and the pre-trained model, which can be formulated as:
% \begin{equation}
% \label{eq:1}
$\delta_t = \mathcal{C}\left(\theta_t - \theta_{base}\right),$
where $\mathcal{C}\left(\cdot\right)$ denotes the delta compression method, $t\in N$, and $\theta_{base}$ is the pretrained model weight $\theta_{pre}$. 
However, as illustrated before, the optimal base model for each task varies.
A single base model may cause severe performance loss on partial tasks and thus cannot achieve optimal performance. 
Therefore, we propose a new paradigm. Instead of a single static base model, we apply a dynamic base model, which can be formulated as:

\begin{equation}
\theta_{base}^t = \theta_{pre} + \lambda_1^t \cdot \tau_{base},
\end{equation}
where base vector $\tau_{base}=\mathcal{C}\left(\frac{\sum_{t=1}^N \left(\theta_{t}-\theta_{pre}\right)}{N}\right)$ is the compressed delta parameter (by BitDelta~\cite{liu2024bitdelta} in our experiments) of the average model weights and $\lambda_1^t$ is a task-specific modulator to adapt the base model to the target task, thus reducing the performance loss.
% Furthermore, the overall scale of the delta parameters can also be adjusted. 
Based on these findings, we propose \method{} (\methodshort{}). We first initialize the shifted base model based on the minimizing L2 distance strategy.
Then, to further boost delta compression's performance, we apply another scaling coefficient $\lambda_2^t$ on the compressed task-specific delta parameter, that is:
\begin{figure}[t]
    \centering
    \includegraphics[width=\linewidth]{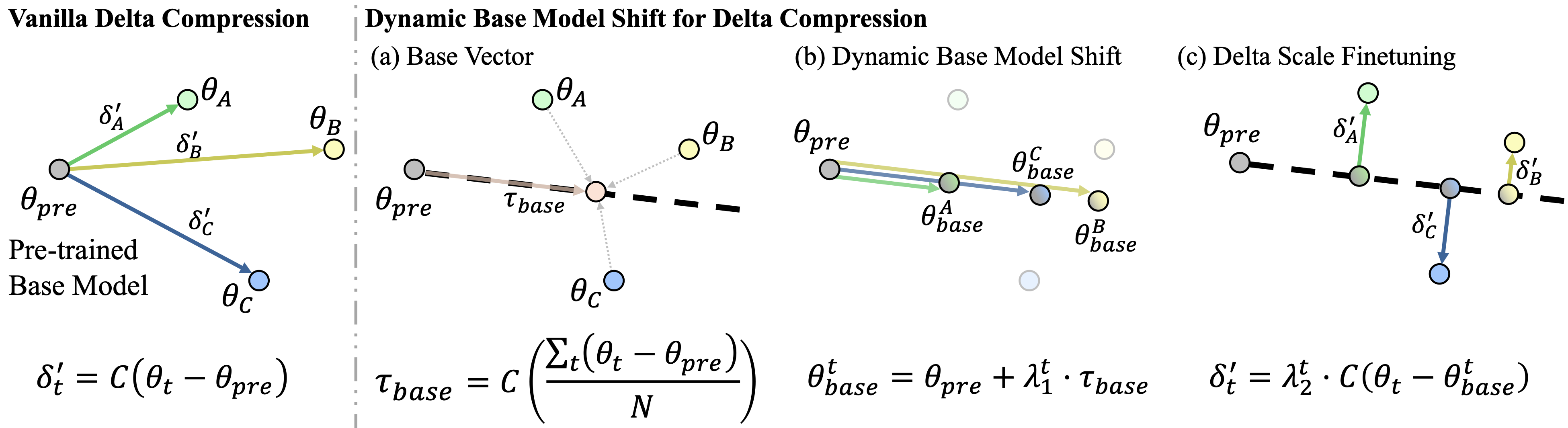}
    \caption{\label{fig:moti} Comparison between vanilla delta compression methods (left) and our \methodshort{} (right), which first (a) obtains a compressed base vector, then we (b) fuse the base vector with the ajusted magnitude to the pretrained model to realize dynamic base model shift. Finally we (c) further dynamically adjust the scales of delta parameters to boost the performance.
    }
    % \vspace{-1cm}
\end{figure}

\begin{equation}
\delta_t' = \lambda_2^t\cdot\mathcal{C}\left(\theta_t-\theta_{base}^t\right).
\end{equation}

To find the optimal $\lambda_1^t$ and $\lambda_2^t$ values for task $t$,
we apply low-cost training on the parameters $\lambda_1^t$ and $\lambda_2^t$. 
During the training process, a partial (10\% in our experiments) unlabeled test dataset is required.
Through training, the performance of delta compression can be effectively improved at a low cost.
The schematic diagram of our \methodshort{} is shown in the right part of Fig.~\ref{fig:moti}.
% The details for training are illustrated in Section~\ref{sec:method}.
% and the comparison of performance and storage overhead can be found in Section~\ref{subsec:storage_overhead}.

In this paper, we demonstrate the effectiveness of our DBMS under the setting of (1)~Language models, including RoBERTa~\cite{liu2019roberta}, GPT-2~\cite{radford2019language}, and Flan-T5~\cite{chung2024scaling, raffel2020exploring} models. The performance improvement is remarkable under all the compression rates and model structure settings, especially when the compression rate is extremely high. For example, our \methodshort{} can improve the performance of GPT-2 models by over $15\%$ under the compression ratio of 99\%.
(2)~Vision transformer~\cite{dosovitskiy2020image} (ViT) models, including ViT-B/32 and ViT-L/14 variants. \methodshort{} can improve the performance of ViT-B/32 and ViT-L/14 models by over $70\%$ and $40\%$ under the compression ratio of $99.8\%$.
(3)~Multi-modal models BEiT3~\cite{beit3}. \methodshort{} shows effectiveness when applied to different kinds of multi-modal tasks.
In short, the proposed \methodshort{} can achieve impressive performance under all the experimental settings, especially under an extremely high compression ratio setting.
More importantly, it can be combined with pruning- and quantization-based delta compression methods.

Our contributions can be summarized as: 

(1) We find that the default paradigm (regarding the pre-trained model as the base model) of existing delta compression methods may not be optimal.
We first investigate this problem and propose \method{}, dubbed \methodshort{}, to shift the base model dynamically, thus solving the performance degradation problem.
(2) We apply two parameters for each task to scale the compressed base vector and the delta parameter. By low-cost fine-tuning of these two parameters, the performance of delta compression can be significantly improved.
(3) \methodshort{} is orthogonal to the existing delta compression method and can be effectively combined with both pruning- and quantization-based methods to boost the performance at low training and storage cost. 
The effectiveness of \methodshort{} is validated through comprehensive experiments on models with different sizes and trained on different modalities, and shows remarkable performance.

\section{Related Work}
\label{sec:related_work}

Delta compression that compresses the delta parameters or task vectors~\cite{Task_Arithmetic}, the difference between the finetuned model parameters and the pre-trained model parameters, can effectively reduce storage and deployment costs.
This approach is especially beneficial in scenarios where a specific model is required for each task.
Technically, delta compression methods can be divided into pruning-based methods and quantization-based methods as follows.

\textbf{Pruning-based methods} eliminate most elements within a task vector randomly~\cite{yu2023language} or based on magnitude~\cite{TIESMerging}.
Ties-Merging~\cite{TIESMerging} first trims 80\% of the task vector values based on magnitude before merging them to reduce weight interference caused by sign conflicts.
EMR-Merging~\cite{huang2024emrmerging} filters elements in task vectors and reserves those elements with the same sign as the elected task vector.
DARE (Drop And REscale)~\cite{yu2023language} is a simple but effective method, which further drops over 90\% of the delta parameters randomly and rescales the remaining elements.
Based on DARE, DAREx~\cite{deng2024dare} re-designs the rescale process to resolve the failing issue of DARE when either the pruning rate or the magnitude of the delta parameters is significant.

\textbf{Quantization-based methods} normally quantize the delta parameters to equivalently 1-bit or even lower~\cite{isik2023gpt, yao2023deltazip}. 
GPT-Zip~\cite{isik2023gpt} first extends GPTQ~\cite{frantar2022gptq} to quantize delta parameters.
BitDelta~\cite{liu2024bitdelta} layer-wise quantizes the delta parameters to 1-bit and finetunes the scale through distillation.
Delta-CoMe~\cite{ping2024deltacome} adopts mixed precision quantization based on the singular values of delta parameters to realize untrained delta compression. 
Delta-DCT~\cite{huang2025seeing} applies the Discrete Cosine Transform (DCT) to delta parameters before applying mixed precision quantization.  
Delta-DQ~\cite{jiang2024deltadq} applies group-wise dropout and separate quantization to the delta parameters.

However, all these methods follow the same paradigm of using the static pre-trained model as the base model for each task, which may not be optimal and cause performance degradation.
In this paper, we in-depth investigate this problem and propose \methodshort{}, which applies an adaptive base model for each task before delta compression and can be combined with various pruning- and quantization-based methods. 
Through training two parameters, which determine the magnitude of the base model shift and the overall scale of delta compression respectively, the performance of delta compression can be significantly improved.

\section{Method}
\label{sec:method}

\begin{figure}[t]
    \centering
    \includegraphics[width=0.985\linewidth]{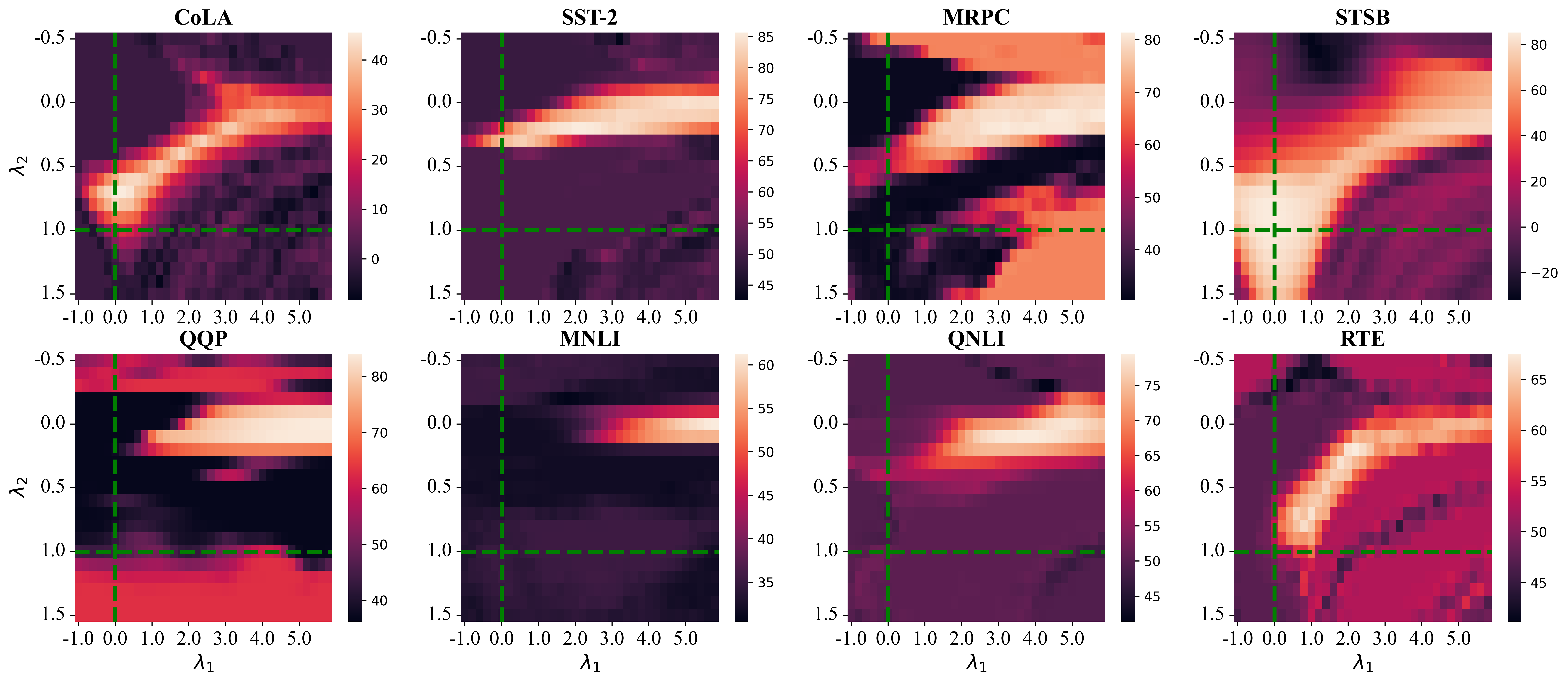}
    \caption{Performance heatmap of compressing RoBERTa models finetuned on the GLUE~\cite{wang2018glue} benchmark under the setting of different $\lambda_1$ and $\lambda_2$ values. We can observe that the original pre-trained base model paradigm ($\lambda_1=0$ and $\lambda_2=1$, which is marked by the intersection of the green dashed lines) rarely achieves optimality, and the optimal values of $\lambda_1$ and $\lambda_2$ vary across different datasets.}
    \label{fig:vis_lambda}
\end{figure}

\subsection{Preliminaries}

Given $N$ tasks $\left[ T_1..T_N \right]$ and their corresponding models $\left[ W_1..W_N \right]$ finetuned from the same pretrained model $W_{pre}$, 
existing delta compression methods directly compress the delta parameters through pruning or quantization:

\begin{equation}
\label{eq:1}
\delta_t = \mathcal{C}\left(W_t - W_{pre}\right),
\end{equation}

where $\mathcal{C}\left(\cdot\right)$ denotes the delta compression algorithm such as DARE~\cite{yu2023language} and BitDelta~\cite{liu2024bitdelta}. It should be noted that in our paper, the BitDelta process does not include the layer-wise scale distillation; that is, we follow the BitDelta-Init setting in \cite{liu2024bitdelta}.
For more details about DARE and BitDelta, please check Appendix~\ref{sec:app_baseline}.

\subsection{\method{}}

\setlength{\intextsep}{2pt}
\setlength{\columnsep}{10pt}
\begin{wrapfigure}{r}{0.525\textwidth}
% \vspace{-5mm}
\begin{minipage}{\linewidth}
\begin{algorithm}[H]

\caption{Algorithm Flow of \methodshort{}} 
\label{algo:algo} 
\KwIn{Finetuned models $W_{1..N}$, pretrained model $W_{pre}$. Partial unlabeled data from each task $X\left[1..N\right]$.}
% \KwOut{Compressed base task vector $\tau_{base}$, scale values $\lambda_1^{[1..N]}$ and $\lambda_2^{[1..N]}$.}

\Comment{\scriptsize Obtain the compressed base vector.}

$\tau_{base} = \mathrm{BitDelta}\left(\frac{\sum_{i=1}^NW_i}{N}-W_{pre}\right)$;

\Comment{\scriptsize Flatten parameters for $\lambda_1$ initialization.}
$\tau_{base}^{flat} = \mathrm{Flatten}\left(\tau_{base}\right)$;
$W_{pre}^{flat} = \mathrm{Flatten}\left(W_{pre}\right)$;

\For{t \textbf{in} $1..N$}{ 
$W_{t}^{flat} = \mathrm{Flatten}\left(W_{t}\right)$

\Comment{\scriptsize Initialize $\lambda_1$ and $\lambda_2$ for task $t$.}

$\lambda_1^t = \frac{\left(W_{t}^{flat}-W_{pre}^{flat}\right)\cdot \tau_{base}^{flat}}{\tau_{base}^{flat}\cdot \tau_{base}^{flat}}$, 
$\lambda_2^t = 1$;

\Comment{\scriptsize Training loop.}

optimizer = Adam([$\lambda_1$, $\lambda_2$],lr=1e-4);

\For{x \textbf{in} $X\left[t\right]$}{
$\tau_t'=\mathcal{C}\left(W_t-\left(W_{pre}+\lambda_1^t\cdot\tau_{base}\right)\right)$;

$W_t'=W_{pre}+\lambda_1^t\cdot\tau_{base}+\lambda_2^t\cdot\tau_t'$;

\Comment{\scriptsize Adam Optimization:}

optimizer.zero\_grad();

% $\min\limits_{\lambda_1,\lambda_2}
$\mathcal{L}=\mathrm{MSE}\left(W_t\left(x\right),W_t'\left(x\right)\right)$;

$\mathcal{L}$.backward(); optimizer.step();
% \Comment{\scriptsize Calculating loss.}

% $\mathcal{L}=\mathrm{MSE}\left(W_t\left(x\right),W_t'\left(x\right)\right)$;

% \Comment{\scriptsize Calculate gradients.}
% $\nabla_{\lambda_1^t} \mathcal{L} = \frac{\partial L}{\partial \lambda_1^t}$;
% $\nabla_{\lambda_2^t} \mathcal{L} = \frac{\partial L}{\partial \lambda_2^t}$;

% \Comment{\scriptsize Update $\lambda_1^t$ and $\lambda_2^t$ using gradient descent}

% $\lambda_1^t = \lambda_1^t - \alpha \cdot \nabla_{\lambda_1^t} \mathcal{L}$; $\lambda_2^t = \lambda_2^t - \alpha \cdot \nabla_{\lambda_2^t} \mathcal{L}$;
}

}
\Return $\tau_{base}$,  $\tau_{1..N}'$, $\lambda_1^{1..N}$, $\lambda_2^{1..N}$.
\end{algorithm}
\end{minipage}
\vspace{-8mm}
\end{wrapfigure}

We believe that using the pre-trained weights $W_{pre}$ is not optimal and may cause significant performance loss. Thus, we apply the dynamic base model, which is the linear interpolation of the pre-trained model and the average merged model, which can be formulated as:

\begin{equation}
\begin{split}
\label{eq:base_model}
\tau_{base}&=\mathrm{BitDelta}\left(W_{avg}-W_{pre}\right),\\
W_{base}^t &= W_{pre} + \lambda_1^t \cdot \tau_{base},
\end{split}
\end{equation}

where $W_{avg} = \frac{\sum_{t=1}^{N}W_t}{N}$. 
Then, we compress the delta parameters based on $W_{base}$.
To further improve the performance, we use another parameter, $\lambda_t^2$, to adjust the overall magnitude of the compressed delta parameters:

\begin{equation}
\delta_t = \lambda_2^t\cdot\mathcal{C}\left(W_t - W_{base}\right),
\end{equation}

where the $\lambda$ values, $\lambda_t^1$ and $\lambda_t^2$ are trainable parameters.

\begin{table}[t]

\centering
\caption{Results of compressing GPT-2 models on seven datasets from GLUE benchmark.}
\label{table:performance_gpt-2} 
\resizebox{0.85\textwidth}{!}{
\begin{tabular}{l|ccccccc|c}
\toprule
Method  & CoLA &SST-2& MRPC& QQP & MNLI&QNLI&RTE& Avg. \\
\midrule
Finetuned & 76.80& 91.17 & 80.39 & 89.64  & 82.00 & 88.27 & 65.34 & 81.94
\\
    \midrule

DARE~\cite{yu2023language}~(95\% Sparse)&
76.61 & 88.76 & 81.61 & 67.18 & 44.76 & 85.17 & 62.82&72.42\\

\midrule
\multicolumn{6}{l@{}}{\textit{w/ \methodshort{}~(Ours)}}\\
\rowcolor{linecolor!45}Init& 76.99& 90.25 & 80.15 & 75.30 & 67.81 & 87.08&64.62 & 77.46\\
\rowcolor{linecolor!45}Post Training&77.18& 90.37 & 80.64 &  87.91 & 79.58 & 87.63 &65.34 & 81.24\\
\midrule

DARE~\cite{yu2023language}~(98\% Sparse)& 76.03 &71.44&81.37& 63.18 & 32.93 & 58.46 & 63.90 & 63.90 \\
\midrule
\multicolumn{6}{l@{}}{\textit{w/ \methodshort{}~(Ours)}} \\
\rowcolor{linecolor!45}Init&76.22 &  86.01 & 79.90&62.93  & 33.17 & 68.22 & 65.34&67.40\\
\rowcolor{linecolor!45}Post Training &76.61 & 88.88 & 80.15 & 85.65& 75.44 & 85.48 &65.34  & 79.65\\
\midrule
DARE~\cite{yu2023language}~(99\% Sparse)&
75.16 & 61.70 & 79.90 & 63.18 & 32.94 & 50.43 & 65.70 & 61.29\\

\midrule
\multicolumn{6}{l@{}}{\textit{w/ \methodshort{}~(Ours)}}\\
\rowcolor{linecolor!45}Init&   71.91 & 73.74 & 79.66 & 63.18 & 32.96  & 50.54 & 65.34 & 62.48\\
\rowcolor{linecolor!45}Post Training&75.93 & 76.15 & 80.64 & 83.87 & 71.81 & 79.84 & 65.70  & 76.27\\
\midrule

% DARE~\cite{yu2023language}~(99.8\% Sparse)&  53.12 & 51.15 & 65.20 & 41.47 & 32.50 & 49.50 & 62.09\\
% \midrule
% \multicolumn{6}{l@{}}{\textit{w/ \methodshort{}~(Ours)}} \\
% \rowcolor{linecolor!45}Init&69.13\\
% \rowcolor{linecolor!45}Post Training & 69.13 & \\
% \midrule

BitDelta~\cite{liu2024bitdelta}& 69.22 & 89.45 & 79.66 & 88.42 & 79.52  &85.81 & 55.60 & 78.24
\\
\midrule
\multicolumn{6}{l@{}}{\textit{w/ \methodshort{}~(Ours)}}\\
\rowcolor{linecolor!45}Init & 75.07 & 90.48&79.90&89.20 & 81.71 & 87.52 & 64.26&81.16\\
\rowcolor{linecolor!45}Post-Training& 77.37&91.28 & 80.64 & 89.56& 82.49 & 88.58 & 65.70&82.23\\

\bottomrule
\end{tabular}
}

\end{table}

\subsubsection{Initialization}
\label{sec:method_init}
For the initialization of $\lambda_1$, we follow the strategy to minimize the L2 distance between the shifted base model and the finetuned model to reduce the performance degradation caused by delta compression methods, that is, for task $t$:
\begin{equation}
\label{eq:obj_ini}
\min_{\lambda_1^t}{\Vert W_{pre}+\lambda_1^t\cdot\tau_{base} - W_t\Vert_2},
\end{equation}

Therefore, the initial values of $\lambda_1$ and $\lambda_2$ could be calculated by:
\begin{equation}
\begin{split}
\left[\lambda_1^t\right]_{init} &= \frac{\left(W_{t}-W_{pre}\right)\cdot \tau_{base}}{\tau_{base}\cdot \tau_{base}},\\
\left[\lambda_2^t\right]_{init} &= 1.
\end{split}
\end{equation}

The detailed theoretical analysis of our initialization strategy is shown in Appendix~\ref{sec_app_theo_init}.
We conduct ablation studies on different initialization strategies of $\lambda_1$ to demonstrate the effectiveness of our strategy. Please check Section~\ref{sec:exp_ablation} for more information.

% $\lambda_2$ is initilized to $1$.
\subsubsection{Training}
To further boost the performance of the proposed \methodshort{}, 
we use the mean square error (MSE) minimization strategy on partial unlabeled test samples.
The objective of the training process is to minimize the MSE between the compressed and uncompressed models' outputs, which is a common practice in knowledge distillation~\cite{gou2021knowledge}.
The optimization objective could be formulated as:
\begin{equation}
% \begin{split}
\min_{\lambda^t_1,\lambda^t_2} \sum_{x\in X_t}\mathrm{MSE}\left( W_t'\left(x\right), W_t\left(x\right)\right),
% \end{split}
\end{equation}
where $W_t'$ can be calculated by:
\begin{equation}
\label{eq:recon}
W_t'=W_{pre}+\lambda_1^t\cdot\tau_{base} +\lambda_2^t\cdot \mathcal{C}\left(W_t-W_{pre}-\lambda_1^t\cdot\tau_{base}\right),
\end{equation}
and $X_t$ is the collection of partial unlabeled test samples sampled in task $T_t$. Specifically, in our experiments, the $X_t$ are 10\% of the input data randomly selected from the test datasets. We use an Adam~\cite{kingma2014adam} optimizer to update the $\lambda_{1}$ and $\lambda_{2}$ values and the learning rate is usually set to $0.0001$.

After the training process, we obtain the BitDelta compressed base delta parameters $\tau_{base}$ applicable to all the tasks, and the compressed task-specific delta parameters $\delta_t=\mathcal{C}\left(W_t-W_{pre}-\lambda_1^t\cdot\tau_{base}\right)$ for task $T_t$, along with two $\lambda^t$ values for scaling.
Before inference on task $T_t$, we obtain the compressed model through Eq.~\ref{eq:recon}.
The effectiveness of our \methodshort{} is theoretically analyzed in Appendix~\ref{sec_app_theo_eff}.

\subsection{Empirical Analysis}
\label{subsec:empirical}

We use a dynamic base model and a controllable scale for each task when applied to RoBERTa~\cite{liu2019roberta} models finetuned on 8 tasks from the GLUE~\cite{wang2018glue} benchmark.
The detailed experimental settings can be found in Section~\ref{sec:exp_roberta}. We test the performance of the compressed models $W_t'$, which can be calculated by Eq.~\ref{eq:recon}, on each task under different values of $\lambda_1$ and $\lambda_2$.
% \begin{equation}
% W_{com}=W_{pre}+\lambda_1^t\cdot\tau_{base} +\lambda_2^t\cdot\mathcal{C}\left(W_t-W_{pre}-\lambda_1^t\cdot\tau_{base}\right)
% \end{equation}
The performance of the compressed model is shown in Fig.~\ref{fig:vis_lambda}.
It can be seen that for each task, the values of $\lambda_1$ and $\lambda_2$ significantly affect the performance. The optimal $\lambda$ values significantly vary for each task. 
Therefore, using partial data to find the optimal $\lambda$ values through training is necessary.
We also provide the visualization results for ViT models and for RoBERTa models under the setting of not compressing $\tau_{base}$ in the Appendix~\ref{sec_app:vis_lambda}.

% the goal of model merging is to obtain a model applicable to all the tasks using finetuned models $\left[ W_1..W_N \right]$ from the same pre-trained model $W_{pre}$ on each task. 
% Existing methods focus on merging the models into a single model $W_M$. Please check Appendix~\ref{sec:app_existing_methods} for detailed information on the existing merging methods. However, a single model can hardly represent all the model weights, thus causing severe performance drops. 
% We discover that the combination of a unified task vector and lightweight task-specific modulators can settle this issue to a significant extent by approximating the task-specific vectors better without any additional tuning. The size of proposed task-specific modulators is discussed in Section~\ref{sec:diff_num_model}, which is much smaller than that of a model.

\section{Experiments}
\label{sec:exp}

\begin{table*}

\centering
\caption{Results of compressing RoBERTa models on eight datasets from GLUE benchmark.}
% \vspace{-8pt}
\label{tab:performance_roberta} 
\resizebox{0.85\textwidth}{!}{
\begin{tabular}{l|cccccccc|c}
\toprule
Methods & CoLA & SST2 & MRPC& STSB & QQP & MNLI & QNLI & RTE & Avg\\
% \multirow{2}{*}{Methods}& \multicolumn{2}{c|}{\small Single-Sentence Tasks}& \multicolumn{3}{c|}{\small Similarity and Paraphrase Tasks} & \multicolumn{3}{c}{\small Inference Tasks} \\%\multirow{2}{*}{Avg Perf}\\
% & \textbf{CoLA} & \textbf{SST2}& \textbf{MRPC} & \textbf{STSB} & \textbf{QQP} & \textbf{MNLI}& \textbf{QNLI} & \textbf{RTE}
%    % \multirow{2}{*}{SUN397}  &  \multirow{2}{*}{Cars}  &  \multirow{2}{*}{RESISC45}  &  \multirow{2}{*}{EuroSAT}  &  \multirow{2}{*}{SVHN}  &  \multirow{2}{*}{GTSRB}  &  \multirow{2}{*}{MNIST}  &  \multirow{2}{*}{DTD}  & \multirow{2}{*}{\textbf{Avg Acc}}
%     \\
% &  Data& inputs & labels & Training&&&&&&&&

% Methods & CoLA & SST2 & MRPC & STSB & QQP & MNLI & QNLI & RTE & Avg Perf 

\midrule
Finetuned & 60.18&	94.04&	89.22&	90.63	&91.41	&87.20	&92.71	&79.06 & 85.56
\\

\midrule

DARE~\cite{yu2023language}~(95\% Sparse)& 61.94 & 93.58 & 87.50& 90.39& 87.56 &74.27 & 91.87 & 76.90 & 83.00 \\
\midrule
\multicolumn{10}{l@{}}{\textit{w/ \methodshort{}~(Ours)}}\\
\rowcolor{linecolor!45}Init& 60.47 &  93.35&88.24 & 90.64 & 87.99&77.77& 91.82 & 77.62&83.49\\
\rowcolor{linecolor!45}Post Training & 60.47 & 93.35 & 88.73 & 90.65 & 88.94&83.91 & 91.38 & 77.62 & 84.38\\
\midrule

DARE~\cite{yu2023language}~(98\% Sparse)&60.99	&92.20&	87.50	&90.06&	63.24&	32.20	&89.97&	78.70&	74.36\\
\midrule
\multicolumn{10}{l@{}}{\textit{w/ \methodshort{}~(Ours)}}\\
\rowcolor{linecolor!45}Init &58.90	&92.55&	87.01&	90.80	&64.39	&31.97	&87.81&	77.62	&		73.88\\
\rowcolor{linecolor!45}Post Training&59.75	&92.55&	87.25	&90.80	&87.38&	75.04&	88.96	&77.62			&82.42\\
\midrule

DARE~\cite{yu2023language}~(99\% Sparse)&59.10&	86.35&	58.82&	89.78&	58.18&	32.71&	81.33&	76.53&	67.85\\
\midrule
\multicolumn{10}{l@{}}{\textit{w/ \methodshort{}~(Ours)}}\\
\rowcolor{linecolor!45}Init&60.08&85.44&84.56&90.14&41.42&33.61&73.09&74.73& 67.88 \\
\rowcolor{linecolor!45}Post Training&60.08&	89.79&	85.29&	90.26	&84.93	&   73.25 &	85.89&	75.45 &80.62 \\
\midrule

DARE~\cite{yu2023language}~(99.8\% Sparse)& 0.00 & 49.89 & 41.91 & 79.07 & 36.88& 33.82 & 50.54 & 63.90 &44.50 \\
\midrule
\multicolumn{10}{l@{}}{\textit{w/ \methodshort{}~(Ours)}} \\
\rowcolor{linecolor!45}Init& 24.37 & 50.92&32.11&84.99&38.47 & 35.45 &48.62 & 50.90 & 45.73\\
\rowcolor{linecolor!45}Post Training&42.98&80.05&74.51& 85.56  & 84.06 &66.69 & 61.36 & 62.46& 69.71 \\
\midrule

BitDelta~\cite{liu2024bitdelta}& 38.91&	93.35	&88.24	&85.97&	90.03	&85.13&90.32&	63.90&	79.48\\
\midrule
\multicolumn{10}{l@{}}{\textit{w/ \methodshort{}~(Ours)}}\\
\rowcolor{linecolor!45}Init& 38.39&	93.35	&88.73&	85.98	&90.98&	87.17&	91.01	&64.26	&		79.98 \\
\rowcolor{linecolor!45}Post-Training& 51.82&	93.92&	89.22	&90.37&	91.24	&87.42	&92.7	&70.03&			83.34\\
\bottomrule
\end{tabular}
}

\end{table*}

We evaluate the performance of (1)~DARE~\cite{yu2023language} under different sparse rates, (2)~BitDelta~\cite{liu2024bitdelta}, and (3)~\methodshort{} combined with both DARE and BitDelta.
All the experiments are conducted on a single NVIDIA A800 GPU.

\begin{table*}[t]
\vspace{-1.5mm}
\centering
\caption{Results of compressing Flan-T5 models on eight datasets from GLUE benchmark.}
% \vspace{-8pt}
\label{tab:performance_t5} 
\resizebox{0.85\textwidth}{!}{
\begin{tabular}{l|cccccccc|c}
\toprule
Methods & CoLA & MNLI & MRPC & QNLI &QQP& RTE&SST2&STSB & Avg\\
% \multirow{2}{*}{Methods}& \multicolumn{2}{c|}{\small Single-Sentence Tasks}& \multicolumn{3}{c|}{\small Similarity and Paraphrase Tasks} & \multicolumn{3}{c}{\small Inference Tasks} \\%\multirow{2}{*}{Avg Perf}\\
% & \textbf{CoLA} & \textbf{SST2}& \textbf{MRPC} & \textbf{STSB} & \textbf{QQP} & \textbf{MNLI}& \textbf{QNLI} & \textbf{RTE}
%    % \multirow{2}{*}{SUN397}  &  \multirow{2}{*}{Cars}  &  \multirow{2}{*}{RESISC45}  &  \multirow{2}{*}{EuroSAT}  &  \multirow{2}{*}{SVHN}  &  \multirow{2}{*}{GTSRB}  &  \multirow{2}{*}{MNIST}  &  \multirow{2}{*}{DTD}  & \multirow{2}{*}{\textbf{Avg Acc}}
%     \\
% &  Data& inputs & labels & Training&&&&&&&&

% Methods & CoLA & SST2 & MRPC & STSB & QQP & MNLI & QNLI & RTE & Avg Perf 

% Pretrained 
\midrule
Finetuned & 74.98	&83.41	&87.50	&91.49&	85.37	&85.92	&93.58&	88.70&		86.37
\\

\midrule

% DARE~\cite{yu2023language}~(95\% Sparse)&
% 74.69 & 82.88 & 86.76 & 91.60 & 85.41 &84.48 & 93.58 & 88.82 & 86.03
% \\
% \midrule
% \multicolumn{10}{l@{}}{\textit{w/ \methodshort{}~(Ours)}}\\
% \rowcolor{linecolor!45}Init& 75.36 & 83.66 & 87.75 &91.43& 85.37 & 85.92 & 93.46 & 87.63 &86.32\\
% \rowcolor{linecolor!45}Post Training \\
% \midrule

% DARE~\cite{yu2023language}~(98\% Sparse)&74.49 & 84.08 &85.05& 91.56&85.13 & 85.20 & 93.46 & 85.93& 85.61\\
% \midrule
% \multicolumn{10}{l@{}}{\textit{w/ \methodshort{}~(Ours)}}\\
% \rowcolor{linecolor!45}Init &75.74 & 83.38 & 86.76 & 91.52 & 85.31 & 85.20 & 93.81 & 86.41&86.02\\
% \rowcolor{linecolor!45}Post Training&\\
% \midrule

DARE~\cite{yu2023language}~(99\% Sparse)&
74.40&	83.44&	87.25&	91.27&	85.27&	84.84&	93.23	&84.91	&	85.58\\

\midrule
\multicolumn{10}{l@{}}{\textit{w/ \methodshort{}~(Ours)}}\\
\rowcolor{linecolor!45}Init& 75.46 &82.90 & 86.03 & 91.78 & 85.39 & 86.64 & 93.69 & 83.05 &85.62 \\
\rowcolor{linecolor!45}Post Training& 75.55 & 83.92 &86.27 & 91.78 & 85.39 & 85.92  & 93.81 & 83.40 & 85.76\\
\midrule

DARE~\cite{yu2023language}~(99.8\% Sparse)& 73.25 & 82.07 & 80.88 & 91.03 & 85.57 & 81.22& 93.35& 84.66& 84.00 \\
\midrule
\multicolumn{10}{l@{}}{\textit{w/ \methodshort{}~(Ours)}} \\
\rowcolor{linecolor!45}Init& 75.07 & 83.21 & 86.27 & 91.36 & 85.27 & 83.39& 93.00 &87.26& 85.60\\
\rowcolor{linecolor!45}Post Training & 75.07 & 83.56 & 86.52 & 91.43 & 85.24 & 83.39 & 93.35 & 87.54&85.76\\
\midrule

BitDelta~\cite{liu2024bitdelta}& 70.09	&83.64	&84.06	&90.94&	85.26&	83.75&	93.46&	86.20&		84.68
\\
\midrule
\multicolumn{10}{l@{}}{\textit{w/ \methodshort{}~(Ours)}}\\
\rowcolor{linecolor!45}Init & 70.09 & 84.09 & 84.56  & 91.16 & 85.28 & 85.20 & 93.46 & 86.83 & 85.08\\
\rowcolor{linecolor!45}Post-Training& 70.28 &84.55 & 84.56  & 91.32 &85.43 & 85.20 & 93.46 & 86.87 & 85.21\\
\bottomrule
\end{tabular}
}
\vspace{-3mm}
\end{table*}

\begin{table*}[t]

\centering
\caption{$\lambda_1$ and $\lambda_2$ values for RoBERTa models before and after training.}
% \vspace{-8pt}
\label{tab:lambda_value_roberta} 
\resizebox{0.75\textwidth}{!}{
\begin{tabular}{ll|cccccccc}
\toprule
\multicolumn{2}{l|}{Methods} & CoLA & SST2 & MRPC& STSB & QQP & MNLI & QNLI & RTE \\
% \multirow{2}{*}{Methods}& \multicolumn{2}{c|}{\small Single-Sentence Tasks}& \multicolumn{3}{c|}{\small Similarity and Paraphrase Tasks} & \multicolumn{3}{c}{\small Inference Tasks} \\%\multirow{2}{*}{Avg Perf}\\
% & \textbf{CoLA} & \textbf{SST2}& \textbf{MRPC} & \textbf{STSB} & \textbf{QQP} & \textbf{MNLI}& \textbf{QNLI} & \textbf{RTE}
%    % \multirow{2}{*}{SUN397}  &  \multirow{2}{*}{Cars}  &  \multirow{2}{*}{RESISC45}  &  \multirow{2}{*}{EuroSAT}  &  \multirow{2}{*}{SVHN}  &  \multirow{2}{*}{GTSRB}  &  \multirow{2}{*}{MNIST}  &  \multirow{2}{*}{DTD}  & \multirow{2}{*}{\textbf{Avg Acc}}
%     \\
% &  Data& inputs & labels & Training&&&&&&&&

% Methods & CoLA & SST2 & MRPC & STSB & QQP & MNLI & QNLI & RTE & Avg Perf 
\midrule
\multicolumn{7}{l}{DARE~\cite{yu2023language} (99.8\% Sparse) w/ Ours}
\\
\midrule
% \rowcolor{linecolor!45}Post Training&42.98&80.05&74.51& 85.56  & 84.06 &66.69 & 61.36 & 62.46 \\
\multirow{2}{*}{Init}
&{$\lambda_1$}
&0.0889 & 0.4930 &0.4429 & 0.0790 & 3.0017 & 3.3504 & 0.5114  &0.0328 
\\

&{$\lambda_2$}
&1.0000 & 1.0000&1.0000 & 1.0000&1.0000 & 1.0000&1.0000 & 1.0000
\\
\midrule
\multirow{2}{*}{Final}
&{$\lambda_1$}
&0.2344 & -0.0561& 0.7013 & 0.1817 & 4.3454 & 3.7923 & 0.3670& 0.2890
\\

&{$\lambda_2$}
&0.8154&0.2968 & 0.3818 & 0.8983 & 0.1490 & 0.1258 &  0.2517& 0.8023
\\
\midrule\multicolumn{7}{l}{BitDelta~\cite{liu2024bitdelta} w/ Ours}
\\
\midrule
\multirow{2}{*}{Init}
&{$\lambda_1$}
&0.0889& 0.4930 & 0.4429 & 0.0790 & 3.0017 & 3.3504 & 0.5114 & 0.0328\\

&{$\lambda_2$}
&1.0000 & 1.0000&1.0000 & 1.0000&1.0000 & 1.0000&1.0000 & 1.0000
\\
\midrule
\multirow{2}{*}{Final}
&{$\lambda_1$}
&0.3913 & 0.7430 &  0.5853 & 0.4161 & 3.9055 &  5.8698&1.0993& 0.1118
\\

&$\lambda_2$
&1.3280 & 1.2500 &  1.1408 & 1.3424&  1.4453&  1.3253& 1.5198& 1.0793

\\

\bottomrule
\end{tabular}
}

\end{table*}

\begin{table*}

\centering
\caption{Results of compressing ViT-B/32 models on eight vision tasks.}

\label{tab:performance_vitbase32} 
\resizebox{0.98\textwidth}{!}{
\begin{tabular}{l|cccccccc|c}
\toprule

{Methods}& SUN397 & Cars &RESISC45 & EuroSAT & SVHN & GTSRB & MNIST & DTD  & \textbf{Avg} \\
\midrule
Finetuned & 75.3  &  77.7  &  96.1  &  99.7  &  97.5  &  98.7  &  99.7  &  79.4 & 90.5   \\
\midrule
DARE~\cite{yu2023language}~(95\% Sparse)&77.5	&74.3	&95.6&	99.7&	97.3	&98.6&	99.7	&77.5	&	90.0  \\
\midrule

\multicolumn{2}{l@{}}{\textit{w/ \methodshort{}~(Ours)}}\\
\rowcolor{linecolor!45}Init&78.2&	75.4	&95.7&	99.7	&97.4	&98.6&	99.6&	78.5	&		90.4
\\
\rowcolor{linecolor!45}Post Training &78.7&	76.4	&95.7	&99.7	&97.4&	98.6	&99.6	&78.8	&		90.6
										\\
\midrule

DARE~\cite{yu2023language}~(98\% Sparse)&72.2&	67.6&	94.9&	99.7&	97.0&	98.4&	99.7&	75.5&		88.1\\
\midrule
% {Traditional MTL} &    73.9  &  74.4  &  93.9  & 98.2    &  95.8  &  98.9   &  99.5   & 77.9 & 88.9  \\
\multicolumn{2}{l@{}}{\textit{w/ \methodshort{}~(Ours)}}\\
\rowcolor{linecolor!45}Init&75.8&70.5&94.8&99.7&97.0&98.5&99.6&76.3&89.0\\
\rowcolor{linecolor!45}Post Training & 78.1&74.5&95.1&99.7&97.0&98.5&99.6&76.4&89.9\\

\midrule
DARE~\cite{yu2023language}~(99\% Sparse)& 50.7&46.1 &91.6& 99.4& 96.1&98.3& 99.6 & 65.5&80.9 \\
\midrule

\multicolumn{2}{l@{}}{\textit{w/ \methodshort{}~(Ours)}}\\

\rowcolor{linecolor!45}Init&68.7&60.2&91.0&99.6&96.4&98.3&99.5&68.2&85.2\\
\rowcolor{linecolor!45}Post Training& 77.2&73.1&93.8&99.7&96.4&98.3&99.6&71.5 & 88.7\\
\midrule
% DARE~\cite{yu2023language}~(99.5\% Sparse)&
% 5.5& 1.1& 56.7  &97.6 & 91.4 & 94.3 & 99.1 & 40.9 & 60.8\\
% \midrule

% \multicolumn{2}{l@{}}{\textit{w/ \methodshort{}~(Ours)}}\\

% \rowcolor{linecolor!45}Init&34.1	&20.6	&50.5	&96.8	&92.5	&91.8	&99.0	&41.3			&65.8\\
% \rowcolor{linecolor!45}Post Training& 76.5	&67.6	&90.4	&98.7	&94.8	&96.4&	99.3&	56.5			&85.0\\
% \midrule
% BitDelta~\cite{liu2024bitdelta}& 78.3 & 75.9 & 95.4& 99.4& 96.4 & 98.2 & 99.2 & 78.4 & 90.1\\
% \midrule
% \multicolumn{2}{l@{}}{\textit{w/ \methodshort{}~(Ours)}}\\
% \rowcolor{linecolor!45}Init & 79.2&78.0&95.4&99.4&96.4&98.2&99.2&78.2 & 90.5\\
% \rowcolor{linecolor!45}Post-Training&79.4&78.3&95.8&99.6&97.2&98.4&99.2&79.0&90.9\\

DARE~\cite{yu2023language}~(99.8\% Sparse)&
0.2&	0.6&	3.5&	35.4&	13.1&	5.0&	10.6&	2.9&	8.9
\\
\midrule

\multicolumn{2}{l@{}}{\textit{w/ \methodshort{}~(Ours)}}\\

\rowcolor{linecolor!45}Init&0.5&	0.5&	3.6&	20.0&	16.7&	6.4&	12.8	&4.3	&		8.1\\
\rowcolor{linecolor!45}Post Training& 75.7&	70.0	&82.8	&96.5	&84.0	&82.4	&98.4&	57.1		&	80.8\\
\midrule
BitDelta~\cite{liu2024bitdelta}& 78.3 & 75.9 & 95.4& 99.4& 96.4 & 98.2 & 99.2 & 78.4 & 90.1\\
\midrule
\multicolumn{2}{l@{}}{\textit{w/ \methodshort{}~(Ours)}}\\
\rowcolor{linecolor!45}Init & 79.2&78.0&95.4&99.4&96.4&98.2&99.2&78.2 & 90.5\\
\rowcolor{linecolor!45}Post-Training&79.4&78.3&95.8&99.6&97.2&98.4&99.2&79.0&90.9\\

\bottomrule
\end{tabular}
}

\end{table*}

\begin{table*}[t]
\vspace{-5mm}
\centering
\caption{Results of compressing ViT-L/14 models on eight vision tasks.}

\label{tab:performance_vitlarge14} 
\resizebox{0.98\textwidth}{!}{
\begin{tabular}{l|cccccccc|c}
\toprule

{Methods}& SUN397 & Cars &RESISC45 & EuroSAT & SVHN & GTSRB & MNIST & DTD  & \textbf{Avg} \\
\midrule
Finetuned & 82.3  &  92.4  &  97.4  &  100  &  98.1  &  99.2  &  99.7  &  84.1  & 94.2    \\
\midrule
DARE~\cite{yu2023language}~(95\% Sparse) &84.4&	91.9&	97.3&	99.7&	98.1&	99.2&	99.7&	84.3 &94.3\\
\midrule

\multicolumn{2}{l@{}}{\textit{w/ \methodshort{}~(Ours)}}\\
\rowcolor{linecolor!45}Init& 84.5	&92.0	&97.3	&99.7	&98.1	&99.2&	99.7&	84.1		&	94.3
\\
\rowcolor{linecolor!45}Post Training& 84.6&	92.3	&97.4&	99.7	&98.1	&99.2&	99.7&	84.3		&	94.4
										\\

\midrule
DARE~\cite{yu2023language}~(98\% Sparse) &83.6&	90.4&	97.0&	99.7&98.0&	99.2&	99.7&	82.9&	93.8\\
\midrule
% {Traditional MTL} &    73.9  &  74.4  &  93.9  & 98.2    &  95.8  &  98.9   &  99.5   & 77.9 & 88.9  \\
\multicolumn{2}{l@{}}{\textit{w/ \methodshort{}~(Ours)}}\\
\rowcolor{linecolor!45}Init&84.1	&91.4	&97.4	&99.7	&98.0	&99.1	&99.7	&83.0	&		94.1\\
\rowcolor{linecolor!45}Post Training &84.5&	91.9&	97.4	&99.7	&98.0&	99.2	&99.7&	83.5&	94.2\\

\midrule
DARE~\cite{yu2023language}~(99\% Sparse)& 81.0&78.5& 96.1& 99.7&97.9& 99.2& 99.7& 78.1& 91.3\\
\midrule

\multicolumn{2}{l@{}}{\textit{w/ \methodshort{}~(Ours)}}\\

\rowcolor{linecolor!45}Init&82.9&	89.2&	96.9&	99.7&	97.9	&99.1	&99.7	&81.8&93.4\\
\rowcolor{linecolor!45}Post Training&84.1	&91.1	&97.2	&99.7&	97.9	&99.1	&99.7	&82.7			&94.0 \\

\midrule

DARE~\cite{yu2023language}~(99.8\% Sparse)&9.2&	2.2&	33.0&	78.8&	87.8&	89.6&	35.5&	29.5&			45.7\\
\midrule
\multicolumn{2}{l@{}}{\textit{w/ \methodshort{}~(Ours)}}\\

\rowcolor{linecolor!45}Init&20.8&	3.2&	37.5&	80.0&	92.1&	92.5&	43.9&	30.2&			50.0\\
\rowcolor{linecolor!45}Post Training& 83.1	&87.5&	92.7	&94.0&	95.2&	98.4&	99.3&	71.4&	90.2		\\

\midrule
BitDelta~\cite{liu2024bitdelta}&84.0 &92.1& 97.2& 99.7& 97.9& 99.0 &99.7& 83.0 &94.1\\
\midrule
\multicolumn{10}{l@{}}{\textit{w/ \methodshort{}~(Ours)}}\\
\rowcolor{linecolor!45}Init & 84.4	&92.4	&97.2&	99.7&	97.9	&99.0&	99.7	&82.9		&	94.2 \\
\rowcolor{linecolor!45}Post-Training & 84.7	&92.6	&97.4	&99.7	&98.1&	99.3	&99.7	&83.8	&		94.4\\

\bottomrule
\end{tabular}
}

\end{table*}

\begin{table*}

\centering
\caption{Results of compressing multi-modal BEiT3 models on vision-language tasks.}
% \vspace{-8pt}
\label{tab:performance_beit} 
\resizebox{0.74\textwidth}{!}{
\begin{tabular}{l|cccc|c}
\toprule
% \multirow{2}{*}{Methods} 
% Methods& Retrieval & Image-Captioning & Image-Classification &  VQA&NLVR  & Avg Perf 
% \\
% \midrule
% & Metric & Accuracy & BLEU4 & Accuracy & 

Methods&\textbf{COCO-Retrieval}  & \textbf{ImageNet-1K} & \textbf{NLVR2}& \textbf{VQAv2} & \textbf{Avg}
\\
% & \multicolumn{1}{|c|}{Metric}& Accuracy($\uparrow$) & BLEU4($\uparrow$) &CIDEr($\uparrow$)&METEOR($\uparrow$) & ROUGE-L($\uparrow$)& Accuracy($\uparrow$) & Accuracy($\uparrow$)  & Accuracy($\uparrow$)
% \\
\midrule

{Individual} & 84.56 &  85.37 & 77.65 &84.39& 82.99 \\
% & 0.8456 & 0.394 &1.337&0.311&0.601& 0.8537 & 0.7765 &0.8439 
\midrule
{DARE~\cite{yu2023language} (60\% Sparse)}
&0.12&50.80&84.07&77.16&53.04\\
\midrule
{\textit{w/ \methodshort{}~(Ours)}}\\
\rowcolor{linecolor!45}{Init} &0.08&70.69 & 84.21 &77.11 & 58.02\\
\rowcolor{linecolor!45}{Post Trainig} &18.64&81.17 & 84.53&77.14&65.37\\
\midrule
{DARE~\cite{yu2023language} (70\% Sparse)}
&0.11&9.67& 84.04 & 76.86 & 42.67\\
\midrule
\multicolumn{5}{l@{}}{\textit{w/ \methodshort{}~(Ours)}}\\
\rowcolor{linecolor!45}{Init} &0.10&32.63 & 83.41&77.11& 48.31 \\
\rowcolor{linecolor!45}{Post Trainig} &16.37&78.59 &83.64 & 77.11& 63.93\\
\midrule
{DARE~\cite{yu2023language} (80\% Sparse)}&0.09&0.71&83.65&76.01&40.12
\\
\midrule
\multicolumn{5}{l@{}}{\textit{w/ \methodshort{}~(Ours)}}\\
\rowcolor{linecolor!45}{Init} &0.12&2.76&83.09&76.14&40.53 \\
\rowcolor{linecolor!45}{Post Trainig} &16.50&74.75&83.22&76.01  &62.62 \\
\midrule

{DARE~\cite{yu2023language} (90\% Sparse)}
&0.11&0.16&82.82&73.51&39.15\\
\midrule
\multicolumn{5}{l@{}}{\textit{w/ \methodshort{}~(Ours)}}\\
\rowcolor{linecolor!45}{Init} &0.12&0.17&80.81 &73.28 & 38.60\\
\rowcolor{linecolor!45}{Post Trainig} &16.38&70.36&81.67&73.84&60.56\\
\midrule
{DARE~\cite{yu2023language} (95\% Sparse)}
&0.12&0.11&76.13&64.02&35.10\\
\midrule
\multicolumn{5}{l@{}}{\textit{w/ \methodshort{}~(Ours)}}\\
\rowcolor{linecolor!45}{Init} &0.07&0.13&73.92 &63.06&34.30\\
\rowcolor{linecolor!45}{Post Trainig} &12.96&61.77&75.00 & 68.04&54.44\\
\midrule
{BitDelta~\cite{liu2024bitdelta}}&0.69  & 79.87& 82.32  & 74.32&59.30
\\
\midrule
\multicolumn{5}{l@{}}{\textit{w/ \methodshort{}~(Ours)}}\\

\rowcolor{linecolor!45}{Init} &25.92&79.39&81.87& 74.52&65.43\\
\rowcolor{linecolor!45}{Post Trainig} &25.34&79.39&82.16&76.31 &  65.80\\
% \midrule 

% \multicolumn{2}{l|}{Weight Averaging} & 1893 & 031	&001	&115	&159 &  6771 & 2800 &  6285
% \\

% \midrule

% \multicolumn{2}{l|}{Task Arithmetic~\cite{Task_Arithmetic}}  & 3177 & 033	&000&	118&	176 & 7081 & 3809 & 6933
% \\

% \multicolumn{2}{l|}{Ties-Merging~\cite{TIESMerging}}& 3929 & 029&	001&	108	&167 & 6978 & 3206& 6717
% \\
% % \midrule

% % \textbf{Task-wise AdaMerging} (\methodshort{}~(Ours))  &58.0 &53.2 &68.8 &85.7 &81.1 &84.4 &92.4  &44.8 &71.1 \\

% % \textbf{Task-wise AdaMerging++} (\methodshort{}~(Ours)) &68 &56.9 &73.1 &83.4 &87.3 &82.4 &95.7  &51 &73.7 \\

% \midrule
% \multicolumn{2}{l|}{(\methodshort{}~(Ours))} & \textbf{7946 }& \textbf{289}& 	\textbf{1.060}&	\textbf{272}&	\textbf{534}& \textbf{7742} & \textbf{7475} & \textbf{7211} \\

\bottomrule
\end{tabular}
}
\vspace{-5mm}

\end{table*}

\subsection{Experiments on language models}
\label{sec:exp_roberta}
\noindent 
\textbf{Settings.}
% We follow the settings from DARE~\cite{yu2023language}.
We validate the performance of RoBERTa-base~\cite{liu2019roberta}, Flan-T5-base~\cite{raffel2020exploring, chung2024scaling} models on different tasks from GLUE~\cite{wang2018glue} benchmark, respectively CoLA~\cite{warstadt2019neural}, 
SST-2~\cite{socher2013recursive}, 
MRPC~\cite{dolan2005automatically},
STS-B~\cite{cer2017semeval},
QQP~\cite{iyer2017first},
MNLI~\cite{williams2017broad},
QNLI~\cite{rajpurkar2016squad}, and RTE~\cite{giampiccolo2007third}.
Following the experimental settings from DARE~\cite{yu2023language},
for RoBERTa models, CoLA is evaluated by the Matthews correlation coefficient, and STS-B is evaluated by the average value of Pearson and Spearman correlation coefficients. The other tasks are evaluated by exact match accuracy.
Following the experimental settings from FusionBench~\cite{tangFusionBenchComprehensiveBenchmark2024},
for Flan-T5 models, STS-B is evaluated by Spearman’s $\rho$, and the other tasks are evaluated by exact match accuracy.
Following the experimental settings from FusionBench~\cite{tangFusionBenchComprehensiveBenchmark2024},
for GPT-2 models, all the tasks are evaluated by exact match accuracy.

\textbf{Results.}
The experimental results for GPT-2, RoBERTa, and Flan-T5 models are shown in Tab.~\ref{table:performance_gpt-2}, \ref{tab:performance_roberta}, and \ref{tab:performance_t5}, respectively. 
It can be seen that delta commpression methods normally results in obvious performance loss, especially under the high compression ratio setting. 
For example, DARE with the sparse rate of $99\%$ causes performance drop of over $20\%$ on GPT-2 models but after combining with our \methodshort{}, this value is sharply reduced to $5\%$.
Similar results are shown on RoBERTa models, where under the sparse rate of $99.8\%$, \methodshort{} can boost the perfromance of DARE by nearly $25\%$.
Additionally, we present the $\lambda_1$ and $\lambda_2$ values before and after training for RoBERTa models in the Tab.~\ref{tab:lambda_value_roberta} when combined with BitDelta and DARE under the sparse rate settings of $99.8\%$.
We provide more results of $\lambda_1$ and $\lambda_2$ values before and after training for GPT-2 models in Appendix~\ref{sec_app:gpt_lambda}.

% The results are shown in \cref{tab:performance_roberta}.
% We observe that \methodshort{} achieves performance comparable to or even exceeding that of finetuned models.
% When compared to BitDelta~\cite{liu2024bitdelta} and Delta-CoMe~\cite{ping2024deltacome}, \methodshort{} outperforms both methods on average scores, with particularly strong performance on specific tasks. For instance, \methodshort{} significantly outperforms other compression methods on CoLA task for the T5-base model. 
% These results suggest that \methodshort{} can achieve promising results on not only LLMs, but also relatively small language models.

% \input{tabs/beit3}

\subsection{Experiments on Vision Transformer models}
\textbf{Settings.} We follow the setting from Task Arithmetic~\cite{Task_Arithmetic} and employ ViT-B/32 and ViT-L/14, two variants of CLIP~\cite{radford2021learning} models' visual encoders, as the pre-trained vision transformer~\cite{dosovitskiy2020image} models. We evaluate the performance on eight image classification tasks, including SUN397~\cite{xiao2010sun}, Cars~\cite{krause20133d}, RESISC45~\cite{cheng2017remote}, EuroSAT~\cite{helber2019eurosat}, SVHN~\cite{yuval2011reading}, GTSRB~\cite{stallkamp2011german}, MNIST~\cite{lecun1998mnist}, and DTD~\cite{cimpoi2014describing}. All the datasets are evaluated by accuracy.

\begin{wrapfigure}{r}{0.455\textwidth}
    \centering
    \includegraphics[width=\linewidth]{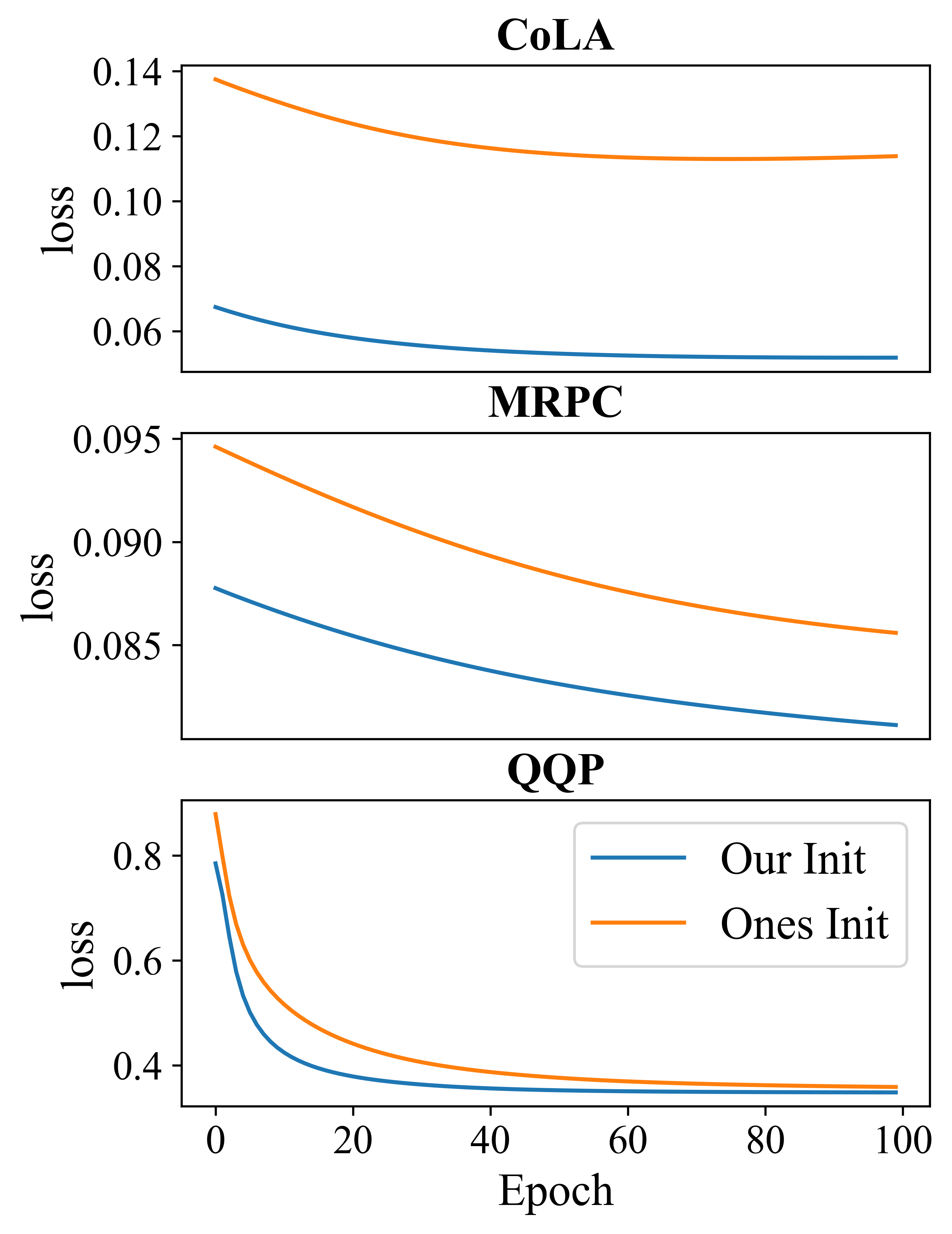}
    \caption{\label{fig:exp_ablation}The loss curve of \methodshort{} using different initaliztion strategies, respectively ours and ones-initialization.}
    \includegraphics[width=\linewidth]{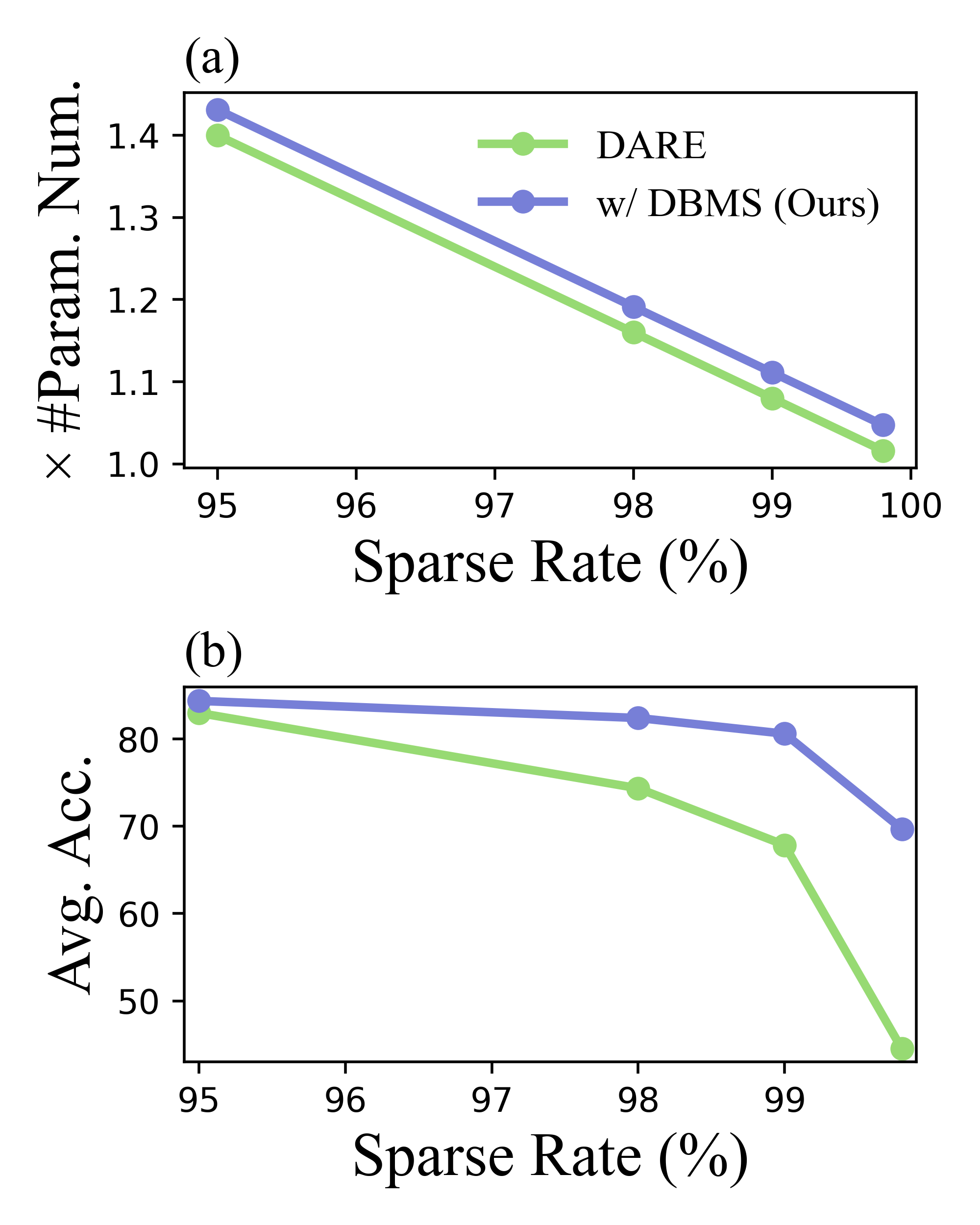}
    \caption{\label{fig:exp_param_and_perf} Comparison of (a) parameter numbers and (b) performance when applied to eight RoBERTa models. }
    \vspace{-2cm}
\end{wrapfigure}

\textbf{Results.} The experimental results for ViT-B/32 and ViT-L/14 models are shown in Tab.~\ref{tab:performance_vitbase32} and \ref{tab:performance_vitlarge14}.
We observe that \methodshort{} performs well in vision models. 
When combined with \methodshort{}, the performance of both DARE~\cite{yu2023language} and BitDelta~\cite{liu2024bitdelta} can significantly improve, especially when the compression rate is extremely high. 
For example, under the sparse rate setting of $99.8\%$, \methodshort{} can improve the average performance of ViT models on eight vision tasks by over $70\%$ and over $40\%$ for ViT-B/32 and ViT-L/14, respectively.
The results demonstrate  the applicability of \methodshort{} to vision models.

% We find that after training, the $\lambda_1$ and $\lambda_2$ values for each task are close to the optimal.

% Furthermore, the average performance of \methodshort{} even surpasses individual finetuned models on both ViT-B/32 models and ViT-L/14 models, while other compression methods and MTL method all experience performance degradation.

% The results of ViT-B/32 models and ViT-L/14 models are respectively shown in Tab.~\ref{tab:performance_vitbase32} and \ref{tab:performance_vitlarge14}.
% We observe that \methodshort{} performs well in vision models, consistently outperforming other compression methods and traditional multi-task learning (MTL) approaches. 
% Furthermore, the average performance of \methodshort{} even surpasses individual finetuned models on both ViT-B/32 models and ViT-L/14 models, while other compression methods and MTL method all experience performance degradation.

% \begin{wrapfigure}{r}{0.4\textwidth}
%     % \vspace{-0.5cm}
%     \centering
%     \includegraphics[width=\linewidth]{figs/loss_comp.png}
%     \caption{\label{fig:exp_ablation}The loss curve of \methodshort{} using different initaliztion strategies, respectively ours and ones-initialization.}
%     \vspace{-1cm}
% \end{wrapfigure}

\subsection{Experiments on multi-modal models}
\textbf{Settings.} We follow the setting from EMR-Merging~\cite{huang2024emrmerging} and employ a BEiT-3~\cite{beit3} model as the pretrained multi-modal model. We select four tasks, respectively ImageNet-1k~\cite{deng2009imagenet} (Image Classification), VQAv2~\cite{goyal2017making} (Visual Question Answering),  NLVR2~\cite{suhr2018corpus} (Visual Reasoning), and COCO Retrieval~\cite{lin2014microsoft} (Image-Text Retrieval). 
All these tasks are evaluated by accuracy.

\textbf{Results.}
The experimental results are shown in Tab.~\ref{tab:performance_beit}.
It can be seen that, compared to vision or language models, BEiT3 models show less redundancy.
However, our \methodshort{} still brings about significant performance improvement under all the compression settings, notably under the setting of sparse rate over 80\%, our \methodshort{} can boost the performance by over 20\%.
The results demonstrate the applicability of our \methodshort{} to multi-modal models.

% \begin{figure}[t]

%     \centering
%     \includegraphics[width=0.33\linewidth]{figs/loss_comp.png}
%     \caption{Ablation. 98\% Sparse}
%     \label{fig:enter-label}
% \end{figure}

% \begin{wrapfigure}{r}{0.4\textwidth}
%     \begin{minipage}[b]{\linewidth}
%     \centering
%     \includegraphics[width=\linewidth]{figs/loss_comp.png}
%     \caption{}
%     \label{fig:enter-label}
%     \end{minipage}
% \end{wrapfigure}

% Ablation on the initialization of $\lambda_1$. (loss figure during training)

\subsection{Ablation Study}
\label{sec:exp_ablation}

As stated in Section~\ref{sec:method_init}, we follow the strategy to minimize the L2 distance between the shifted base model and the finetuned model.
To demonstrate the effectiveness of our initialization strategy, we conduct ablation studies on the initialization of $\lambda_1$.
We conduct experiments on RoBERTa models and apply different initialization strategies, respectively ours and \textit{ones initialization}, also referred to as \textit{Ones Init}, which means initializing $\lambda_1$ and $\lambda_2$ to $1.0$.
We record the loss curve of three different tasks during the training process of the first $100$ epochs, as shown in Fig.~\ref{fig:exp_ablation}.
It can be seen that using our initialization strategy could accelerate the convergence of \methodshort{}, thus showing the effectiveness of our initialization strategy.

\subsection{Discussion: Storage Overhead}
\label{subsec:storage_overhead}
Compared to vanilla DARE~\cite{yu2023language} and BitDelta~\cite{liu2024bitdelta}, our \methodshort{} has some additional storage overhead, which is $\tau_{base}$ in Eq.~\ref{eq:recon}.
Due to the reason that it is compressed to 1-bit through BitDelta~\cite{liu2024bitdelta}, the storage overhead is not obvious.
In Fig.~\ref{fig:exp_param_and_perf}, we compared the storage costs and performance of \methodshort{} when applied to RoBERTa models on eight tasks when combined with DARE with different compression rates.
It can be seen that our \methodshort{} requires only slightly more storage costs compared to DARE while showing a significant performance gain, especially when the compression rate is over 99\%, demonstrating the parameter efficiency of the proposed \methodshort{}.

% compared to layer-wise quantization, the proposed \methodshort{} has storage overhead. 
% We present the overhead of our \methodshort{} by comparing the storage costs of \methodshort{}, individual finetuned models, a merged multi-task model, and existing delta compression methods, under the settings of different task numbers and different patch size $p$ in \cref{fig:overhead}(a) and \cref{fig:overhead}(b), respectively.
% We assume that all the methods are applied to 7B LLMs.
% It can be seen that the storage costs of the proposed method are slightly greater than those of the existing methods but still significantly fewer than individual finetuned models.

% \begin{wrapfigure}{r}{0.52\textwidth}
%     \centering
%     \includegraphics[width=\linewidth]{figs/loss_comp.png}
%     % \vspace{-0.15cm}
%     \captionsetup{type=figure}
%     \caption{\label{fig:perf_compare}This is a fig.}
% \end{wrapfigure}
\section{Conclusion}
\label{sec:conclusion}

In this paper, we investigate the impact of different base models on the performance of delta compression, including both delta pruning and delta quantization. 
Based on the finding that a single static base model may cause unavoidable performance loss, we propose \method{} (\methodshort{}), to adapt the base model to the target task dynamically and adjust the overall scale of delta parameters, thus reducing performance loss caused by delta compression.
Through low-cost training of two parameters, \methodshort{} can effectively improve the performance, especially under the setting of an extremely high compression ratio.
The effectiveness of \methodshort{} is validated by comprehensive experiments on various benchmarks under vision, language, and multi-modal settings.

% study tuning-free and high-performance model merging. We first attribute the severe performance degradation of existing merging methods to that a single model can hardly simulate all the models' performance. Then we propose \method{} (\methodshort{}), which does not require any data access or additional training for tuning.
% % which merges model weights into a unified model and task-specific modules without any tuning requirement of any data access or additional training. 
% % We validate the performance of \methodshort{} under multiple settings and \methodshort{} outperforms other alternatives.
% The effectiveness of \methodshort{} is validated by comprehensive experiments on various classical benchmarks and newly established benchmarks under vision, NLP, PEFT, and multi-modal settings.

% In this work, we investigate the distinctive redundancy mech-
% anisms of upcycled MoE models, and introduce an innova-
% tive DeRS paradigm that remodels MoE experts into one
% shared base weight and multiple exclusive compact weights.
% Further, we propose DeRS compression and DeRS upcycling
% to enhance the efficiency of expert parameters in upcycled
% MoE models during inference and training, respectively. We
% conduct comprehensive experiments to support the effective-
% ness of our proposals. Future works could focus on improv-
% ing the parameter-efficient techniques in DeRS compression
% and DeRS upcycling or extending the DeRS paradigm to
% scenarios with higher training budgets.
\section{Acknowledgement}
\label{sec:acknowledgement}
This work is supported by Shanghai Natural Science Foundation (No. 23ZR1402900), National Natural Science Foundation of China (No. 62071127 and No. 62306261),  National Key Research and Development Program of China (No. 2022ZD0160101).
The computations in this research were performed using the CFFF platform of Fudan University.

{
\small
\bibliographystyle{abbrv}
\bibliography{main}
}
\newpage
\appendix
\vspace{20pt}

\textbf{{\Large Appendix for \method{}}}

\section{Theoretical Analysis}
\label{sec_app_theo}
\subsection{Initialization Strategy}
\label{sec_app_theo_init}
\methodshort{} applies L2 distance minimization strategy to initialize $\lambda_1^t$ and $\lambda_2^t$. For task $t$, the objective is shown in Eq.~\ref{eq:obj_ini} and can be rewritten as:

\begin{small}
\begin{equation}
\begin{split}
&\min_{\lambda_1^t}{\Vert W_{pre}+\lambda_1^t\cdot\tau_{base} - W_t\Vert_2^2}\\
=&\min_{\lambda_1^t}{\left[\left( W_{pre}+\lambda_1^t\cdot\tau_{base} - W_t\right)^T\left( W_{pre}+\lambda_1^t\cdot\tau_{base} - W_t\right)\right]}\\
=&\min_{\lambda_1^t}{\left[\left(\lambda_1^t\right)^2\tau_{base}^T\tau_{base}+\lambda_1^t\tau_{base}^T\left(W_{pre}-W_t\right)+\left(W_{pre}-W_t\right)^T\left(W_{pre}-W_t\right)+\lambda_1^t\left(W_{pre}-W_t\right)^T\tau_{base}\right]}
\end{split}
\end{equation}
\end{small}

We assume that:
\begin{small}
\begin{equation}
f\left(\lambda_1^t\right)=\left(\lambda_1^t\right)^2\tau_{base}^T\tau_{base}+\lambda_1^t\tau_{base}^T\left(W_{pre}-W_t\right)+\left(W_{pre}-W_t\right)^T\left(W_{pre}-W_t\right)+\lambda_1^t\left(W_{pre}-W_t\right)^T\tau_{base} 
\end{equation}
\end{small}  

To obtain the minimal value of $f\left(\lambda_1^t\right)$, 
we set the first derivative of $f\left(\lambda_1^t\right)$ with respect to $\lambda_1^t$ to $0$. That is:

\begin{equation}
\begin{split}
\frac{\mathrm{d}f\left(\lambda_1^t\right)}{\mathrm{d}\lambda_1^t}=&2\lambda_1^t\tau_{base}^T\tau_{base}+\tau_{base}^T\left(W_{pre}-W_t\right)+\left(W_{pre}-W_t\right)^T\tau_{base}\\
=&0
\end{split}
\end{equation}

Solve this function, and thus the initial value of $\lambda_1^t$ can be calculated by:

\begin{equation}
\left[\lambda_1^t\right]_{init} = \frac{\left(W_{t}-W_{pre}\right)\cdot \tau_{base}}{\tau_{base}\cdot \tau_{base}}.
\end{equation}

\subsection{Effectiveness Analysis of \method{}}
\label{sec_app_theo_eff}
The delta parameters of the original paradigm for compressing $W_t$ can be formulated by:
\begin{equation}
\label{eq:wtori}
\Delta_t^{ori}=W_t-W_{pre},
\end{equation}
while after applying our \methodshort{}, the delta parameters can be formulated by:
\begin{equation}
\Delta_t^{ours}=W_t-W_{pre}-\lambda_1^t\tau_{base},
\end{equation}
Eq.~\ref{eq:wtori} could be rewritten as:
\begin{equation}
\Delta_t^{ori}=\Delta_t^{ours} + \lambda_1^t\tau_{base}.
\end{equation}

From the initialization conditions of $\lambda_1^t$, that is minimizing the L2 distance between $\lambda_1^t\tau_{base}$ and $\Delta_t^{ours}$, we can know that 
$\lambda_1^t\tau_{base}$ and $\Delta_t^{ours}$ are orthogonal, $\lambda_1^t\tau_{base}\cdot\Delta_t^{ours}=0$.
Therefore, the variance of $\Delta_t^{ori}$ could be written as:
\begin{equation}
\mathrm{Var}\left(\Delta_t^{ori}\right) = \mathrm{Var}\left(\Delta_t^{ours}\right) + \left(\lambda_1^t\right)^2\mathrm{Var}\left(\tau_{base}\right),
\end{equation}

We can obtain that $\mathrm{Var}\left(\Delta_t^{ori}\right) > \mathrm{Var}\left(\Delta_t^{ours}\right)$.
That is, after applying our \methodshort{}, the variance of delta parameters for compressing is lowered, making it easier to compress.
This explains the effectiveness of our \methodshort{} theoretically.

\begin{table*}[t]
\centering
\caption{$\lambda_1$ and $\lambda_2$ values for GPT-2 models before and after training.}
% \vspace{-8pt}
\label{tab:gpt_value_lambda} 
\resizebox{0.8\textwidth}{!}{
\begin{tabular}{ll|ccccccc}
\toprule
\multicolumn{2}{l|}{Methods}  & CoLA &SST-2& MRPC& QQP & MNLI&QNLI&RTE \\
\midrule
\multicolumn{7}{l}{DARE~\cite{yu2023language} (99\% Sparse) w/ Ours}
\\
\midrule
% \rowcolor{linecolor!45}Post Training&42.98&80.05&74.51& 85.56  & 84.06 &66.69 & 61.36 & 62.46 \\
\multirow{2}{*}{Init}
&{$\lambda_1$}
 & 0.0898 & 0.4923&0.0440 & 2.6556  & 2.8466 & 0.8298 & 0.0321
% &0.0889 & 0.4930 &0.4429 & 0.0790 & 3.0017 & 3.3504 & 0.5114  &0.0328 
\\

&{$\lambda_2$}
&1.0000 & 1.0000&1.0000 & 1.0000&1.0000 & 1.0000&1.0000 
\\
\midrule
\multirow{2}{*}{Final}
&{$\lambda_1$}
&-0.0516 & 0.3650 &  -0.0071 & 1.9235 & 1.8578 & 1.4430&-0.0443
\\

&{$\lambda_2$}
 & 0.8450 &  0.8802 & 0.9574  &  0.3554 & 0.2634 & 0.6052 & 0.9392
\\
\midrule\multicolumn{7}{l}{BitDelta~\cite{liu2024bitdelta} w/ Ours}
\\
\midrule
\multirow{2}{*}{Init}
&{$\lambda_1$}
&0.0898 & 0.4923 &0.0440 & 2.6556 & 2.8466 & 0.8398 &0.0321
\\
&{$\lambda_2$}
&1.0000 & 1.0000&1.0000 & 1.0000&1.0000 & 1.0000&1.0000 
\\
\midrule
\multirow{2}{*}{Final}
&{$\lambda_1$} & 0.4058 & 0.7084 & 0.1640 & 5.0290 & 6.9434 & 1.1650 & 0.1389
\\

& $\lambda_2$ & 1.3225&  1.2208 & 1.1147 & 1.4564 &1.3647  & 1.4147  & 1.1074
\\

\bottomrule
\end{tabular}
}

\end{table*}

\begin{figure}[t]
    \centering
    \includegraphics[width=\linewidth]{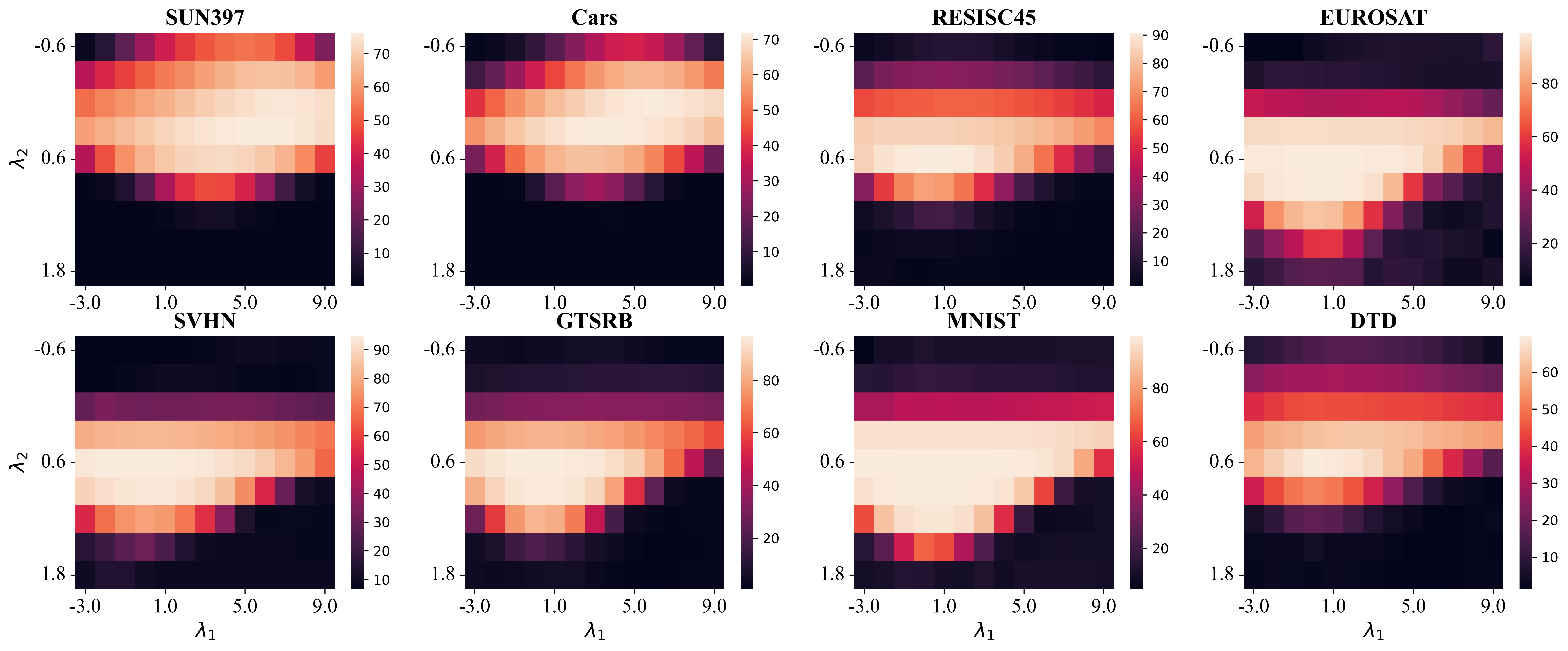}
    \caption{Performance heatmap of compressing ViT-B/32 models finetuned on eight vision tasks under the setting of different $\lambda_1$ and $\lambda_2$ values.}
    \label{fig:vit-b}
\end{figure}

% \section{Algorithm flow of \methodshort{}}
% \label{sec:app_algo}
\section{Baseline Methods}
\label{sec:app_baseline}
\textbf{DARE}~\cite{yu2023language}~(\textbf{D}rop \textbf{A}nd \textbf{RE}scale)
applies a random mask to the delta matrix $\Delta W$.
Under the sparse rate of $p\in\left(0,1\right)$, mask $m$ can be obtained by:

\begin{equation}
m \sim \mathrm{Bernoulli}\left(p\right),
\end{equation}

Then the compressed delta matrix can be calculated through:

\begin{equation}
\hat{\Delta W} = \frac{1}{1-p}\left( m\odot\Delta W\right),
\end{equation}

where $\odot$ denotes element-wise multiplication.

\textbf{BitDelta}~\cite{liu2024bitdelta} obtains a binarized estimator by encoding the sign bits of the delta matrix $\Delta W$:

\begin{equation}
\hat{\Delta W} = \alpha \odot \mathrm{Sign}\left(\Delta W\right),
\end{equation}

where $\alpha$ is the scaling factor, the average of the absolute values of $\Delta W$.
That is, for $\Delta W \in \mathbb{R}^{m\times n}$
$\alpha=\frac{\mathrm{sum}\left(\vert \Delta W\vert\right)}{n\cdot m}$.
$\mathrm{Sign}\left(\cdot\right)$ obtains the sign matrix by $\mathrm{Sign}\left(\Delta W\right)=\mathrm{where}(\Delta W>0, +1, -1)$, which is a 1-bit matrix.
\section{$\lambda$ values for GPT-2 models}
\label{sec_app:gpt_lambda}
In Tab.~\ref{tab:gpt_value_lambda}, we provide the $\lambda_1$ and $\lambda_2$ values before and after training for GPT-2 models when combined with BitDelta and DARE under the sparse rate settings of $99\%$.

\begin{figure}[t]
    \centering
    \includegraphics[width=\linewidth]{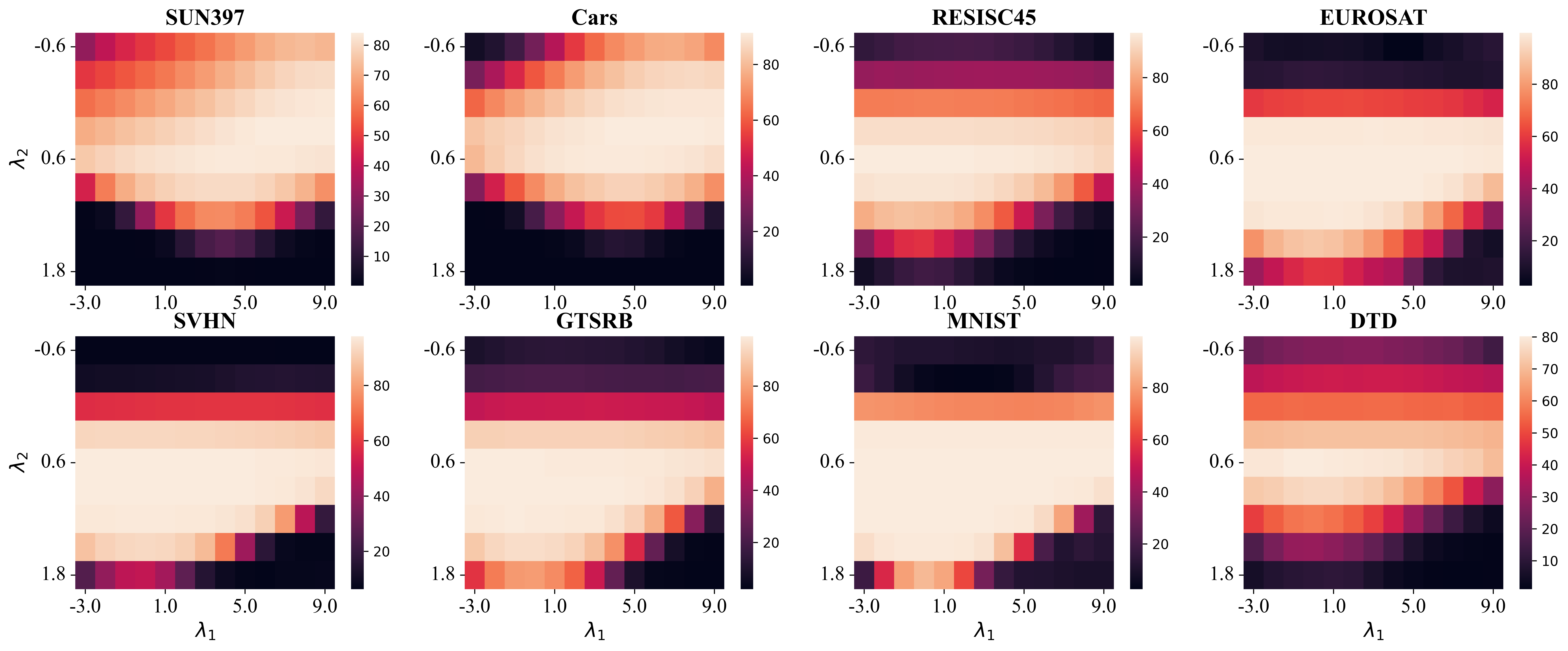}
    \caption{Performance heatmap of compressing ViT-L/14 models finetuned on eight vision tasks under the setting of different $\lambda_1$ and $\lambda_2$ values.}
    \label{fig:vit-l}
\end{figure}

\begin{figure}[t]
    \centering
    \includegraphics[width=\linewidth]{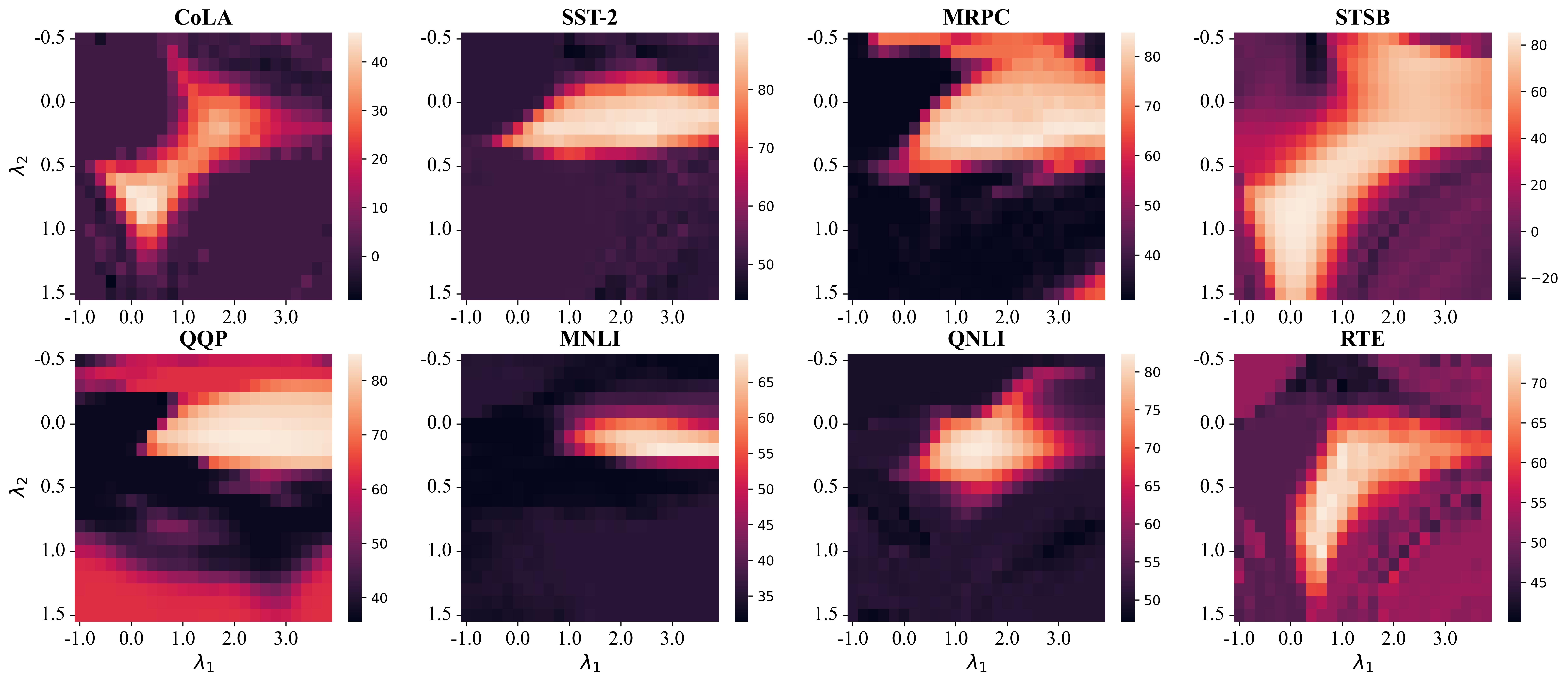}
    \caption{Performance heatmap of compressing RoBERTa models finetuned on the GLUE~\cite{wang2018glue} benchmark under the setting of different $\lambda_1$ and $\lambda_2$ values. Note that in this figure, the model is compressed through Eq.~\ref{eq:recon_no_compress_base}, which does not include base task vector compression. }
    \label{fig:roberta_no_bd}
\end{figure}

\section{More visualization results}
\label{sec_app:vis_lambda}
In Section~\ref{subsec:empirical}, we visualize the results for compressing RoBERTa models. Here we provide more results.
In Fig.~\ref{fig:vit-b} and \ref{fig:vit-l}, we visualize the results for ViT-B/32 and ViT-L/14 models under different $\lambda_1$ and $\lambda_2$ settings in Eq.~\ref{eq:recon}.
In Fig.~\ref{fig:roberta_no_bd}, we show the results for RoBERTa models following the setting of not compressing $\tau_{base}$, that is:

\begin{equation}
\begin{split}
\label{eq:recon_no_compress_base}
\tau_{base}&=\theta_{avg}-\theta_{pre},\\
W_t'&=W_{pre}+\lambda_1^t\cdot\tau_{base} +\lambda_2^t\cdot \mathcal{C}\left(W_t-W_{pre}-\lambda_1^t\cdot\tau_{base}\right).
\end{split}
\end{equation}

% \section{$\lambda_1$ and $\lambda_2$ values presentation}

\section{Limitations and future works}
\label{sec:limitations}
Despite the convincing results, the proposed method suffers from several limitations. 
On the one hand, when combined with existing delta compression methods, \methodshort{} requires a little additional memory to store the shared base task vector, i.e., $\tau_{base}$ in Eq.~\ref{eq:base_model}, as illustrated in Section~\ref{subsec:storage_overhead}.
On the other hand, \methodshort{} requires partial unlabeled data and some additional training of $\lambda_1$ and $\lambda_2$ for each task, which brings about some computational costs.

Further improving the performance and even exploring the upper bound performance while reducing the storage and training costs of delta compression are significant directions for our future work.

% Further improving the performance of delta compression and exploring the performance upper bound while lowering the compression costs is a significant direction for future work.

% \input{tabs/lambda_value_roberta}

\end{document}